\documentclass[10pt,twocolumn,letterpaper]{article}

\usepackage[pagenumbers]{cvpr} % To force page numbers, e.g. for an arXiv version

\definecolor{cvprblue}{rgb}{0.21,0.49,0.74}
\usepackage[pagebackref,breaklinks,colorlinks,allcolors=cvprblue]{hyperref}

\def\paperID{39213} % *** Enter the Paper ID here
\def\confName{CVPR}
\def\confYear{2026}

\title{GeoDiff4D: Geometry-Aware Diffusion for 4D Head Avatar Reconstruction}

\author{
    Chao Xu$^{1, \dagger}$ \quad
    Xiaochen Zhao$^{1}$ \quad
    Xiang Deng$^{1}$ \quad
    Jingxiang Sun$^{1}$ \quad    
    Donglin Di$^{2}$ \quad
    Zhuo Su$^{1, \ddagger}$ \quad
    Yebin Liu$^{1,\S}$\\
    $^1$Tsinghua University \quad $^2$Li Auto
}

\usepackage{booktabs}
\usepackage{multirow}
\usepackage{graphicx}
\usepackage{float}

\begin{document}

\twocolumn[{%
\renewcommand\twocolumn[1][]{#1}%
\maketitle

\begin{center}
    \centering
    \captionsetup{type=figure}
    \includegraphics[width=\textwidth]{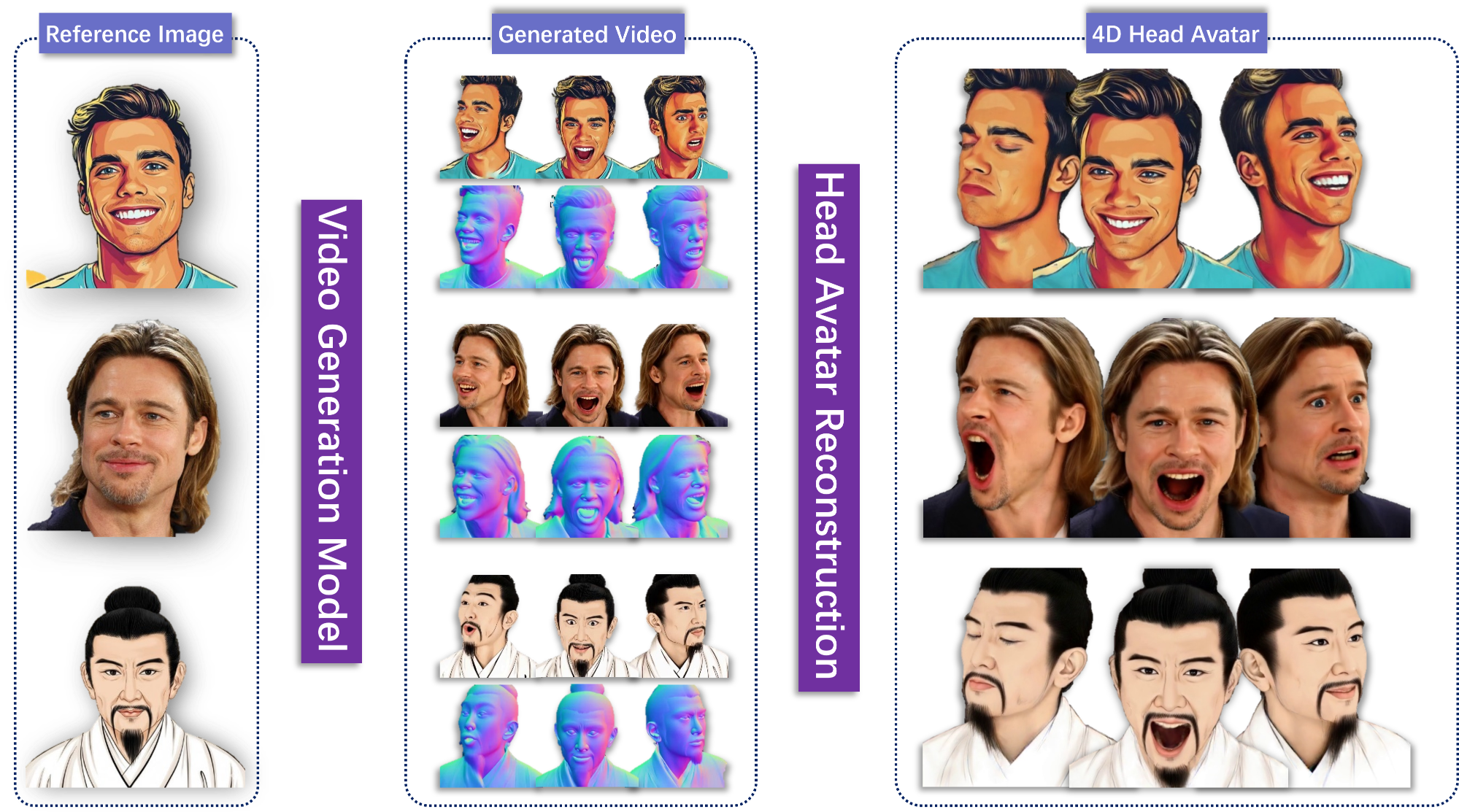}
    \captionof{figure}{We present GeoDiff4D, a framework that reconstructs animatable 4D head avatars from a single portrait image through geometry-aware diffusion. By jointly predicting portrait image frames and surface normals with a pose-free expression encoder, our method trains 3D Gaussians under dual supervision, achieving exceptional identity preservation and 3D consistency.}
    \label{fig:teaser}
\end{center}%
}]

\renewcommand{\thefootnote}{\fnsymbol{footnote}}
\footnotetext[2]{Intern. \quad $^{\ddagger}$Project leader. \quad $^{\S}$Corresponding author.}

\begin{abstract}
Reconstructing photorealistic and animatable 4D head avatars from a single portrait image remains a fundamental challenge in computer vision. While diffusion models have enabled remarkable progress in image and video generation for avatar reconstruction, existing methods primarily rely on 2D priors and struggle to achieve consistent 3D geometry. We propose a novel framework that leverages geometry-aware diffusion to learn strong geometry priors for high-fidelity head avatar reconstruction. Our approach jointly synthesizes portrait images and corresponding surface normals, while a pose-free expression encoder captures implicit expression representations. Both synthesized images and expression latents are incorporated into 3D Gaussian-based avatars, enabling photorealistic rendering with accurate geometry. Extensive experiments demonstrate that our method substantially outperforms state-of-the-art approaches in visual quality, expression fidelity, and cross-identity generalization, while supporting real-time rendering. Our project page: https://lyxcc127.github.io/geodiff4d/

\end{abstract}
\section{Introduction}
\label{sec:intro}
% introduce challenge, combine 2d & 3d, cite
The reconstruction of animatable and expressive head avatars from a single portrait image has attracted considerable attention in computer vision and computer graphics, given its wide range of applications in education, filmmaking, gaming, teleconferencing, and virtual reality. However, as a highly under-constrained problem, the main challenge in achieving photorealistic and faithful avatar reconstruction lies in effectively learning strong priors from large-scale datasets to enhance both identity preservation and subtle facial motion fidelity. In this paper, we present a general framework that learns geometry priors from both multi-view and monocular datasets to reconstruct high-fidelity, photorealistic 4D head avatars from a single portrait image.

Some recent methods~\cite{zhao2025x,luo2025dreamactorm1holisticexpressiverobust,deng2024portrait4dv2}, benefiting from the impressive capabilities of diffusion models, have achieved high-quality and vivid 2D portrait animation. These approaches typically excel at preserving identity and transferring expressions. However, their inherent lack of 3D consistency often leads to significant quality degradation under novel views.
Other methods~\cite{chu2024gagavatar,he2025lam,chu2024gpavatargeneralizableprecisehead} introduce explicit 3D representations to improve geometric consistency, but at the cost of reduced identity preservation and limited ability to capture and reproduce subtle facial expressions. More recently, several approaches~\cite{taubner2025cap4d,taubner2025mvp4d} have combined diffusion-based generative models with 3D Gaussian Splatting (3DGS)~\cite{kerbl3Dgaussians}, synthesizing person-specific data with diverse expressions and head poses via diffusion models to facilitate head avatar reconstruction. However, three key challenges still remain: (1) existing methods often rely on facial landmarks, implicit motion representations, or 3D Morphable Model (3DMM) parameters for expression control, which often struggle to achieve both 3D consistency and expressiveness simultaneously; (2) the diffusion models primarily learn 2D priors, such as pixel-level correspondences, without fully capturing the underlying 3D geometry; (3) these methods generally supervise the avatar reconstruction only with the generated portrait images, resulting in a weak connection between diffusion and reconstruction stages, thus failing to fully exploit the diffusion models' capability of knowledge distillation.

To overcome these challenges, we propose GeoDiff4D, a geometry-aware diffusion framework for 4D avatar reconstruction (Fig.~\ref{fig:teaser}). First, we recognize that expressive yet geometrically consistent expression representations are crucial for guiding the diffusion model to generate vivid and multi-view consistent results. To this end, we introduce a pose-free expression encoder that extracts implicit expression latents capable of capturing subtle facial details. We enhance its 3D consistency through explicit head-pose control and a cross-view pairing strategy during training, enabling the encoder to maintain geometric coherence across different viewpoints while preserving expressiveness. Second, we develop a joint image–normal diffusion model that predicts portrait images together with their corresponding surface normals. By modeling their joint distribution, the diffusion model learns to leverage the rich 3D geometric cues encoded in normals—information absent in RGB appearance alone—thereby significantly enhancing geometric consistency. Finally, we optimize the avatar model using the generated images, normals, and expression latents together, effectively transferring the geometric priors from the diffusion model to enable high-quality 4D reconstruction.

% view consistent exp encoder
% joint representation
% gs
Our contributions can be summarized as follows:
\begin{itemize}
\item We present the first video generation model that jointly synthesizes portrait frames and surface normals, enabling the diffusion model to learn 3D-aware priors rather than purely 2D appearance statistics.
\item We propose a pose-free expression encoder trained with cross-view pairing that captures expressive facial dynamics while ensuring robust view consistency.
\item We propose a head avatar model that leverages the generated images, normals, and expression latents to jointly optimize 3D Gaussian Splatting.
\end{itemize}
    
\section{Related Work}
\label{sec:relatedwork}

\subsection{Head Avatar Reconstruction}

\begin{figure*}[ht]
  \centering
  \includegraphics[width=\linewidth]{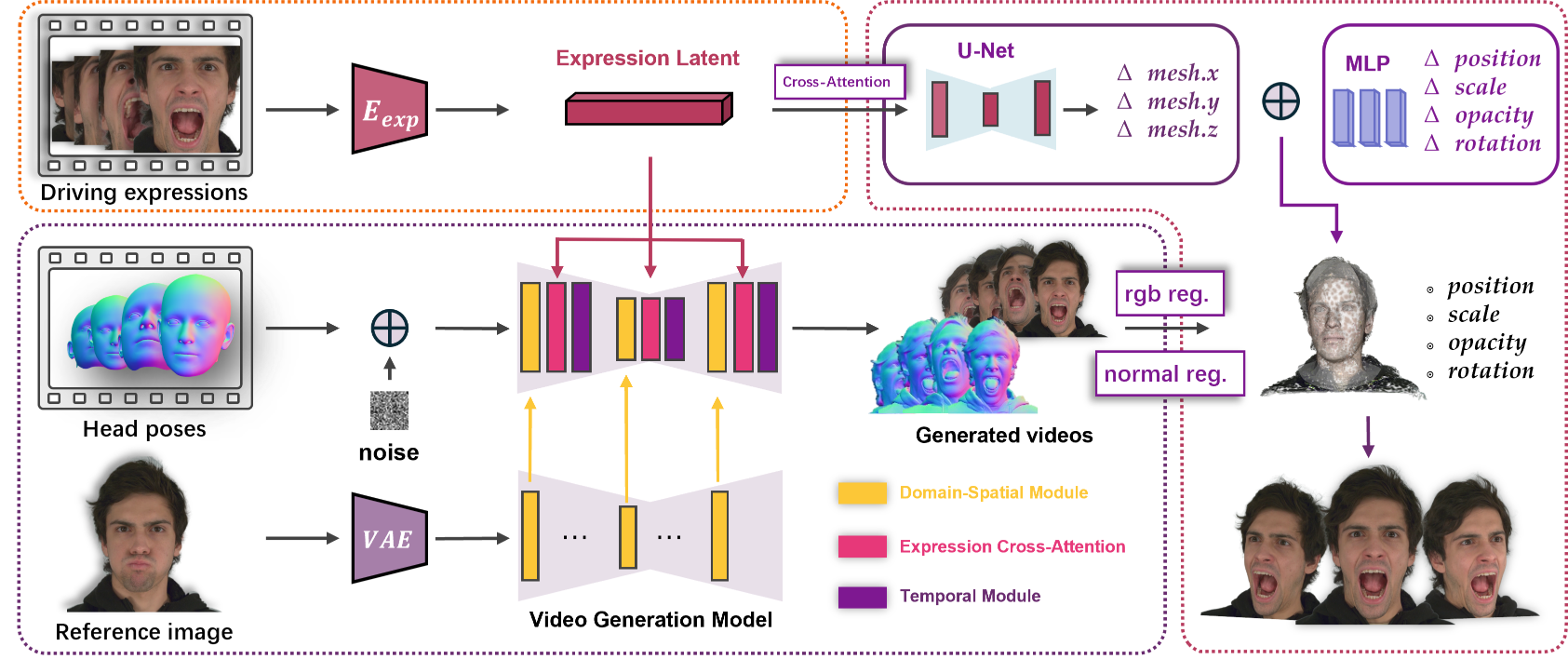}
  \caption{Overall architecture. Our system takes a reference image, driving expressions, and head poses as input. Specifically, the reference image is encoded into hierarchical identity embeddings using a pretrained VAE and UNet-based reference network. Driving expressions are compressed into low-dimensional latents via a pose-free expression encoder. Both embeddings are injected into the diffusion model through cross-attention, while head pose maps concatenated with noise serve as inputs. The model then jointly predicts portrait images and surface normals. For 3D reconstruction, a UNet refines FLAME meshes using expression latents through cross-attention, and an MLP captures Gaussian dynamics. Finally, the generated surface normals provide additional geometric supervision that further enhances the reconstruction fidelity.}
  \label{fig:architecture}
\end{figure*}

Previous methods~\cite{grassal2021neural, ma2021pixelcodecavatars,Zheng2023pointavatar} leverage 3D Morphable Models (3DMMs)~\cite{FLAME:SiggraphAsia2017, wang2022faceverse, gerig2017morphablefacemodels} to reconstruct human heads with explicit control over facial expressions and head poses, enabling efficient animation and rendering. However, their performance is limited by the inherent fixed topology of mesh structures.
Other studies have integrated Neural Radiance Fields (NeRFs)~\cite{mildenhall2020nerf} with parametric head models to achieve view-consistent and photorealistic reconstructions~\cite{zielonka2023instantvolumetricheadavatars, zheng2022imavatar, Gafni_2021_CVPR, athar2022rignerf}. While these approaches effectively capture intricate details, they often suffer from slow rendering speed and long training time.
Recent works have adopted 3D Gaussian Splatting (3DGS)~\cite{kerbl3Dgaussians}, which represents scenes using explicit 3D Gaussian primitives and enables real-time rendering through efficient tile-based rasterization. Some approaches leverage synchronized camera arrays to achieve photorealistic, high-fidelity reconstruction~\cite{qian2024gaussianavatars, lee2025surfhead, xu2023gaussianheadavatar}. Other methods require only a monocular video as input, providing a more practical alternative~\cite{chen2024monogaussianavatar, shao2024splattingavatar, xiang2024flashavatar, HRAvatar}, albeit with a trade-off between data requirements and view consistency. However, these methods often struggle to produce competitive results when the dataset lacks diversity in head poses and facial expressions.

More recently, 3D avatar reconstruction from a single or a few portrait images has attracted significant attention. Existing methods can be broadly categorized into two groups. The first group employs generalizable frameworks~\cite{guo2025segadrivable3dgaussian, he2025lam, chu2024gagavatar, zheng2025headgapfewshot3dhead}, which achieve generalization across different identities by training on large-scale portrait datasets. These models learn strong priors that map identity embeddings to specific 3D representations and typically animate avatars using 3DMM parameters. The second group follows a two-stage pipeline~\cite{taubner2025cap4d, taubner2025mvp4d}, leveraging powerful portrait generation models to synthesize diverse portrait frames from a given image, which are then used to optimize a head avatar.

\subsection{Portrait Animation}
Previous methods primarily focus on GANs~\cite{goodfellow2014generativeadversarialnetworks}, which synthesize facial expressions and head poses by warping and rendering source images. Some approaches incorporate explicit control signals, such as facial landmarks or 3DMM parameters, to improve control and disentanglement between motion and appearance~\cite{gao2023highfidelityfreelycontrollabletalking, siarohin2020ordermotionmodelimage, wang2021oneshotfreeviewneuraltalkinghead, guo2024liveportrait,chan2021piganperiodicimplicitgenerative,chan2022efficientgeometryaware3dgenerative,sun2023next3d}, but they often struggle to capture fine-grained facial details. Other methods leverage implicit representations to reproduce subtle facial expressions~\cite{drobyshev2023megaportraitsoneshotmegapixelneural, drobyshev2024emoportraits, xu2024vasa1lifelikeaudiodriventalking, wang2022progressivedisentangledrepresentationlearning, deng2024portrait4dv2,tran2023voodoo}, yet they are prone to identity leakage. Moreover, GAN-based approaches generally face challenges in generating high-quality outputs and handling out-of-domain portraits due to the inherent limitations of the architecture.
Recently, Latent Diffusion Models~\cite{rombach2021highresolution} have demonstrated strong generative capabilities in portrait synthesis. Several works explore frameworks that combine dual U-Nets~\cite{cao2023masactrltuningfreemutualselfattention} with plug-and-play motion modules~\cite{guo2024animatediffanimatepersonalizedtexttoimage} to achieve temporally coherent motion and consistent appearance across frames~\cite{tian2024emoemoteportraitalive, xie2024xportraitexpressiveportraitanimation}. Other approaches introduce implicit expression representation to capture expressive facial motion and achieve fantastic animation results~\cite{luo2025dreamactorm1holisticexpressiverobust,zhao2025x}.

\subsection{Joint Representation}
Recent studies have explored combining joint representations with diffusion models across multiple tasks. Some approaches employ generative diffusion models to simultaneously estimate geometric attributes such as depth and surface normals from single images of static scenes~\cite{fu2024geowizardunleashingdiffusionpriors, zhang2023jointnetextendingtexttoimagediffusion}. Others focus on jointly generating color images and corresponding modalities of 3D assets or human bodies, such as motion maps or normal maps, by introducing cross-domain attention mechanisms~\cite{long2023wonder3dsingleimage3d, chefer2025videojam}.
Collectively, these works demonstrate that incorporating joint representations enables models to become geometry-aware, thereby enhancing the quality and structural coherence of generated results. Inspired by these findings, we introduce surface normals paired with portrait frames as joint representation for human portrait synthesis to improve the 3D consistency of the generated color images, while the predicted surface normals further serve as supervision signals for optimizing avatar reconstruction.
\section{Method}
\label{sec:method}
GeoDiff4D consists of three key components. First, a pose-free expression encoder extracts a 1D view-consistent expression latent from a single image, capturing facial dynamics while disentangling head pose. Second, a video generation model conditioned on the reference image and expression latent jointly synthesizes a sequence of portrait frames and surface normals. Finally, an animatable 4D avatar is reconstructed via 3D Gaussian Splatting, leveraging the generated images, normals, and expression latents to achieve high-fidelity animation. The full architecture of GeoDiff4D is illustrated in Figure~\ref{fig:architecture}.

\subsection{Pose-Free Expression Encoder}
\label{subsec:expencoder}
We employ an implicit representation to capture fine-grained facial details and propose a cross-view pairing training strategy that encourages the encoder to extract consistent expression representations across different viewpoints. Through joint optimization with the diffusion model, both components learn complementary and view-consistent features. Experiments further demonstrate the robustness and cross-view consistency of our encoder (See Section.~\ref{experiment:expencoder}).

\paragraph{Implicit expression representation.}
Following prior work~\cite{zhao2025x}, we use an expression encoder $E_\text{mot}$ to encode an image into a low-dimensional latent $f_\text{mot}$ that captures facial details while discarding spatial appearance information, encouraging disentanglement between expression and identity. Instead of combining head pose and expression into a single latent, we disentangle head pose from the implicit representation by introducing an explicit head pose control signal (See Section.~\ref{para:headposecontrol}). The resulting expression latent then guides the diffusion model via cross-attention to achieve more accurate and expressive synthesis.

\begin{figure}[ht]
  \centering
  \includegraphics[width=\linewidth]{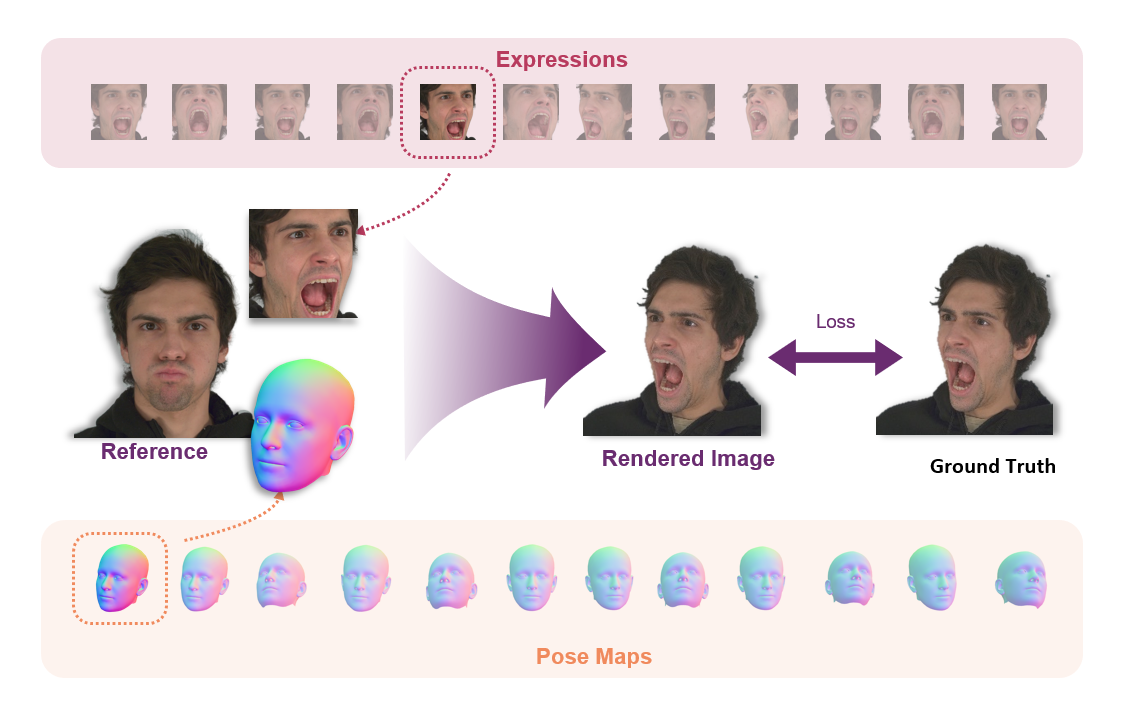}
  \caption{Cross-view pairing training strategy. For each identity and timestep, frames from different viewpoints are paired with consistent expressions but varying poses, enabling the encoder to learn view-invariant representations.}
  \label{fig:trainingstrategy}
\end{figure}

\paragraph{Cross-View Pairing Training Strategy.}
To leverage the multi-view nature of our datasets, we introduce a cross-view pairing training strategy for learning robust, view-invariant expression representations. We sample frames of the same identity and timestep from different viewpoints to form cross-view pairs, where driving and target frames share identical expressions but differ in viewpoint (Figure~\ref{fig:trainingstrategy}). This effectively mitigates head pose and identity leakage, encouraging the model to focus on expression-specific features. The strategy facilitates learning of a pose-free expression encoder and enhances 3D awareness in the diffusion model.

\paragraph{Loss and Augmentation.}
The encoder is trained end-to-end with the diffusion model, supervised solely by the diffusion denoising loss without any auxiliary specific objective. To further promote spatial invariance in the learned embeddings, we apply data augmentation exclusively to the cropped facial driving images, including pixelwise augmentations (brightness, contrast, saturation, noise, and blur) to simulate diverse capture conditions, and spatial augmentations (scaling, rotation, and translation) to reduce sensitivity to spatial layout.

\subsection{Geometry-Aware Video Generation Model}
\label{subsec:vgm}
We adapt a UNet-based latent diffusion framework to jointly generate portrait frames and their corresponding surface normals while achieving disentangled control over facial expression and head pose. Furthermore, we introduce a synthetic dataset with ground-truth surface normal annotations to enhance both the generalizability of the model and the quality of generated surface normals.

\paragraph{Architecture.}
Following X-NeMo~\cite{zhao2025x}, we adopt a UNet-based latent diffusion framework that integrates a reference network $R$, temporal modules and an expression encoder $E$. We adapt this framework for predicting joint distribution of portrait images and surface normals. Specifically, a pretrained auto-encoder $V$ compresses the reference image $I_{ref}$ into low-dimension latent codes for computational efficiency. The reference network then extracts hierarchical identity features $F_{ref}$ from these latents. Simultaneously, the pose-free expression encoder $E$ (See~\ref{subsec:expencoder}) extracts implicit expression representations $F_{exp}=\{f_{exp}\}_{n=1}^{N}$ from the driving frames $I_{exp}=\{i_{exp}\}_{n=1}^{N}$. A sequence of driving normal maps (See Section.~\ref{para:headposecontrol}) $M_{drv}=\{m_{drv}\}_{n=1}^{N}$ are interpolated and concatenated with the latents, and the diffusion model jointly denoises the noisy RGB and normal latents, progressively generating $Z_{rgb}=\{z_{rgb}\}_{n=1}^{N}$ and $Z_{norm}=\{z_{norm}\}_{n=1}^{N}$. The model is conditioned on $F_{ref}$ and $F_{exp}$ via cross-attention mechanisms. Finally, the auto-encoder decodes $Z_{rgb}$ and $Z_{norm}$ back to RGB space, yielding the synthesized frames $I_{rgb}=\{i_{rgb}\}_{n=1}^{N}$ and $I_{norm}=\{i_{norm}\}_{n=1}^{N}$.
In summary, the model learns the joint distribution:
\begin{equation}
    P(I_{rgb},I_{norm}|I_{ref},M_{ref},I_{exp},M_{drv})
\end{equation}

\paragraph{Joint Appearance-Normal Representations.}
\label{para:jointrep}
To enhance 3D awareness, we introduce surface normals as an additional denoising target and modify the diffusion model to jointly generate RGB images and its corresponding normal maps. To obtain normals, we use an off-the-shelf normal estimator~\cite{saleh2025david}, which predicts accurate and detailed results with rich facial structures such as wrinkles and hair strands.

During training, we separately encode the target video clips and their corresponding normal clips to obtain latent representations $Z$ with shape $[B\times D\times C\times T\times H\times W]$. The latents from different domains are then concatenated along the domain dimension $D$ where $C$ and $T$ denote the channel and temporal dimensions, respectively. To ensure consistency across domains, identical noise is applied to the latents at each timestep, and class label embedding is introduced to distinguish domain-specific latents. To enable effective cross-domain interactions, we replace vanilla 2D self-attention with 3D Domain-Spatial attention modules, allowing elements from both domains to effectively learn inter-domain relationships. Specifically, the domain and batch dimensions are merged $[(B \times D) \; C \times T \times H \times W]$ for convolutional processing and separated again before the attention modules. Latents from different domains are then concatenated along the width dimension $W$, forming a tensor of shape $[B\times C\times T\times H\times (2W)]$ for attention. This design preserves domain independence in convolutional layers while allowing controlled information exchange.

% Moreover, the strong alignment between the generated RGB images and their corresponding normal maps allows the normals to be directly used as supervision for Gaussian Reconstruction(See \ref{subsec:gsavatar}).

\paragraph{Head Pose Map Conditioning.}
\label{para:headposecontrol}
To enable explicit head pose control, we introduce head pose maps $M_{tar}=\{m_{tar}^{n}\}_{n=1}^{N}$ which are essentially normal maps that indicate target head poses during both training and inference. We utilize an off-the-shelf FLAME tracker~\cite{qian2024vhap} to extract 3D Morphable Models (3DMM) parameters including shape, expression, translation, and rotation, which can be used to reconstruct a corresponding mesh of a personalized 3D mesh. The head pose maps are then generated by rasterizing the vertex normals of the mesh onto the 2D image plane using the corresponding camera intrinsics and extrinsics. To minimize the identity leakage, we set expression related parameters to zero when rendering the normal maps. To ensure spatial compatibility with the noisy latent input, we interpolate the head pose maps $M\in R^{B\times (V/F)\times C\times H_{m}\times W_{m}}$ to match the latent resolution $Z\in R^{B\times (V/F)\times C\times H_{z}\times W_{z}}$ , where $V$ and $F$ denotes view dimension and temporal dimension. The interpolated head pose maps are then concatenated with the latent features before feeding them into the reference model and denoising model.

\paragraph{Synthetic Dataset.}
Current public datasets with real human portraits lack high-quality surface normal annotations, and pseudo normals obtained from off-the-shelf estimators are inherently limited by estimator accuracy. Synthetic datasets, by contrast, provide precise ground-truth normals and have proven effective for various face-related tasks. We thus incorporate SynthHuman~\cite{saleh2025david}, which offers diverse identities with ground-truth normals, to supplement our multi-view training data, enhancing model generalizability and generated normal quality.

To effectively combine synthetic and multi-view datasets during joint training, we adopt a weighted random sampling strategy with manually-set importance weights adjusted by sample count. This balances dataset diversity while ensuring that multi-view data is sampled approximately 10× more frequently per sample than synthetic data, reflecting its greater volume and reliability, while smaller synthetic datasets remain adequately represented.

\subsection{4D Reconstruction}
\label{subsec:gsavatar}
Our 4D avatar reconstruction builds upon the approach of GaussianAvatars~\cite{qian2024gaussianavatars}, which is based on 3D Gaussian Splatting~\cite{kerbl3Dgaussians} and learns a set of 3D Gaussian primitives attached to the triangles of the FLAME mesh. Specifically, given a reference image, an expression-driving video, and head pose maps, we first generate a sequence of portrait frames along with their associated surface normals. We then optimize a set of 3D Gaussian primitives under the supervision of the generated portraits and surface normals. Furthermore, a hierarchical refinement strategy is employed to enhance geometric consistency and appearance fidelity.

\paragraph{Hierarchical Refinements.}
We treat the generated portrait video as a monocular input and employ the off-the-shelf tracker Pixel3DMM~\cite{giebenhain2025pixel3dmm} to estimate the initial FLAME parameters, including shape, expression, and head pose. Attaching the triangles to the FLAME mesh forms a two-fold pipeline: the FLAME parameters drive the FLAME mesh, and the FLAME mesh in turn animates the 3D Gaussians. However, monocular FLAME tracking remains a challenging problem, and the inherent discrepancy between the 3DMM template and real human facial geometry often degrades the quality of 3DGS optimization. To mitigate this issue, we introduce a hierarchical refinement strategy that progressively captures fine-grained facial details and improves overall geometric fidelity.

First, we introduce learnable residuals for FLAME parameters to compensate for tracking errors. Second, following prior work~\cite{taubner2025cap4d}, we remesh the FLAME head and employ a U-Net to predict per-vertex deformations. Instead of reorganizing a UV mesh by sampling adjacent pixels in the UV map, we construct a face graph in 3D space and reorganize the UV mesh according to original spatial relationships by querying this graph. Furthermore, unlike previous methods that define deformation maps in world space, we use a position map of the FLAME mesh animated solely by expression-related parameters in canonical space and condition the U-Net on the expression latent from our pose-free expression encoder (See~\ref{subsec:expencoder}) via cross-attention. Finally, we employ a lightweight MLP to predict per-primitive Gaussian attribute residuals, since shared attributes across expressions cannot fully capture expression-dependent dynamics.

\paragraph{Surface Normal Regularization.}
Smooth and accurate surface normals are essential for reducing rendering artifacts in novel-view synthesis. While prior methods benefit from multi-view frames with broad facial coverage and diverse head poses, they still struggle to recover high-quality normals. To address this limitation, we introduce pseudo normals predicted by our video generation model as strong supervisory signals for Gaussian Splatting.
Following GaussianShader~\cite{jiang2023gaussianshader}, we take the shortest axis of each Gaussian primitive as its normal $\hat{n}$, extend the Gaussian Splatting representation with dedicated normal channels, and attach $\hat{n}$ to the color features during rendering. We then regularize the rendered normals using a $L_1$ loss over the foreground region:
\begin{equation}
    \mathcal{L}_n = \lambda_n\mathcal{L}_1(\hat{n},\alpha n)
\end{equation}
where $\hat{n}$ and $n$ denote the predicted and pseudo ground-truth normals, $\alpha$ is the foreground mask, $\lambda_n$ is the loss weight and $\mathcal{L}_1$ is the $L_1$ Loss.
\section{Implementation}
\subsection{Video Generation Model}
We train our model using a combination of multi-view datasets~\cite{pan2024renderme, kirschstein2023nersemble} and synthetic datasets~\cite{saleh2025david}. All data are processed to 512×512 resolution, including color images, surface normals, and foreground masks. Additionally, we employ an off-the-shelf 3DMM tracker~\cite{qian2024vhap} to estimate 3DMM parameters and render head pose maps for each frame.
Training follows a two-stage pipeline. In the first stage, the model is trained without temporal modules with a batch size of 32. In the second stage, 16-frame sequences are incorporated for temporal learning with a batch size of 8, with padding to maintain consistent sequence lengths. Both stages use AdamW~\cite{yao2021adahessianadaptivesecondorder} with a learning rate of $1e-5$, trained for 80K and 20K iterations respectively on 4 A800 GPUs, taking approximately 3-4 days in total. All three datasets are used in both stages. Further discussion of computational efficiency is provided in the supplementary material.

\subsection{Head Avatar Reconstruction}
For data generation, we use our video generation model to synthesize portrait videos with 12 views and approximately 200 frames, using a 25-step DDIM~\cite{song2022denoisingdiffusionimplicitmodels} schedule, taking ~1 hour on a single NVIDIA H800 GPU. Head avatar reconstruction follows GaussianAvatars~\cite{qian2024gaussianavatars} training, running 100K steps in 3 hours on an RTX 3090.

\section{Experiments}
\label{sec:experiments}

\begin{figure*}[ht]
  \centering
  \includegraphics[width=\linewidth]{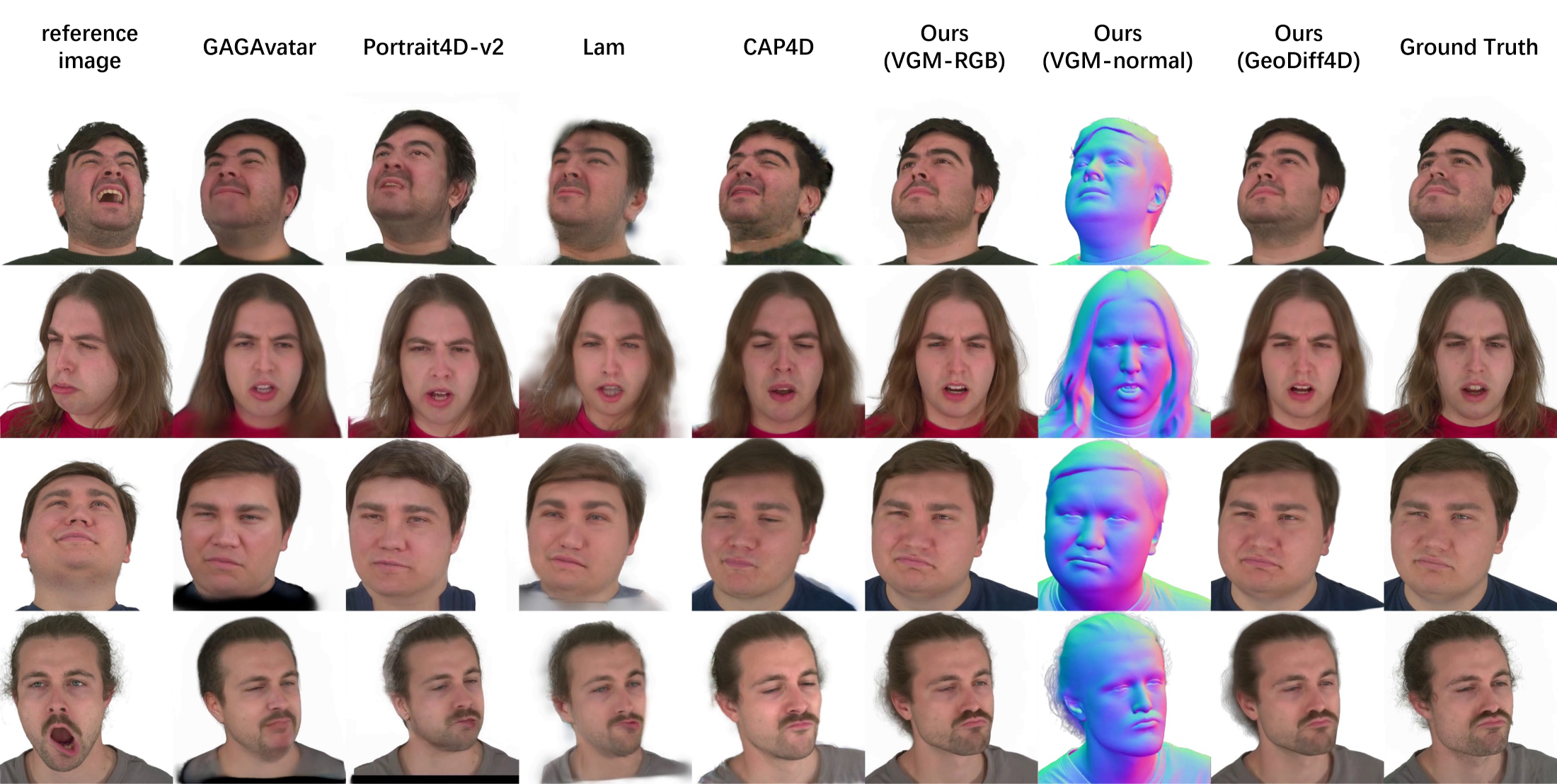}
  \caption{Self-Reenactment results. We conduct this experiment on a subset of the NeRSemblev2 dataset containing a large number of extreme head poses in both the reference images and driving sequences, enabling a comprehensive evaluation of model performance. We also show surface normals generated by our video generation model.}
  \label{fig:selfreenact}
\end{figure*}

\subsection{Experiment Setting}
\paragraph{Metrics.}
We employ three image-based metrics to evaluate the photometric quality of generated frames: PSNR, SSIM~\cite{1284395}, and LPIPS~\cite{zhang2018unreasonableeffectivenessdeepfeatures}, assessing reconstruction fidelity and perceptual similarity. Temporal coherence is evaluated using JOD~\cite{mantiuk2021fovvideovdp}, while identity preservation is measured via the cosine similarity of identity embeddings (CSIM)~\cite{Deng_2022}. To quantify the accuracy of head pose and expression transfer, we further report Average Keypoint Distance (AKD) and Average Expression Distance (AED).

\paragraph{Baselines.}
We compare our method with several single-view 4D head avatar reconstruction systems: Portrait4D-v2~\cite{deng2024portrait4dv2}, GAGAvatar~\cite{chu2024gagavatar}, LAM~\cite{he2025lam}, and CAP4D~\cite{taubner2025cap4d} (P3DMM~\cite{giebenhain2025pixel3dmm} version). These methods represent recent advances encompassing both feed-forward and optimization-based paradigms, providing a comprehensive benchmark for evaluating realism, expression accuracy, and generalization. Comparisons with more methods (X-NeMo~\cite{zhao2025x}, Wan-Animate~\cite{wan2025}, LivePortrait~\cite{guo2024liveportrait} and VoodooXP~\cite{tran2024voodooxpexpressiveoneshot}) are provided in the supplementary material.

\subsection{Self-Reenactment}
\label{sec:seflreenact}
We evaluate self-reenactment on ten unseen subjects from NeRSemblev2~\cite{kirschstein2023nersemble}, using 2 sequences per subject with 4 of 16 camera views, yielding 80 driving clips. Reference images are sampled from the same sequences but different viewpoints, ensuring comprehensive evaluation across diverse head poses and expressions.

As shown in Table~\ref{table:metrics}, our VGM achieves the best performance across all image quality metrics, demonstrating superior fidelity, identity preservation, and temporal consistency. GeoDiff4D ranks second on most metrics, outperforming all baselines. Figure~\ref{fig:selfreenact} shows our method excels in image quality, identity similarity, and expression transfer. Additionally, VGM generates normal maps with rich head details and high consistency with RGB outputs.

\begin{table}[ht]
\centering
\small % Add this to reduce font size
\resizebox{\columnwidth}{!}{%
\begin{tabular}{@{}l*{5}{c}|*{2}{c}@{}}
\toprule
\multirow{2}{*}{\textbf{Method}} & 
\multicolumn{5}{c|}{\textbf{Self-reenactment}} & 
\multicolumn{2}{c}{\textbf{Cross-reenactment}} \\
\cmidrule(lr){2-6} \cmidrule(lr){7-8}
 & PSNR $\uparrow$ & SSIM $\uparrow$ & LPIPS $\downarrow$ & CSIM $\uparrow$ & JOD $\uparrow$ & CSIM $\uparrow$ & JOD $\uparrow$ \\
\midrule
GAGAvatar~\cite{chu2024gagavatar} & 17.550 & 0.789 & 0.229 & 0.714 & 6.244 & 0.588 & \underline{5.081} \\
Portrait4D-v2~\cite{deng2024portrait4dv2} & 13.689 & 0.701 & 0.310 & 0.702 & 4.933 & 0.608 & 4.656 \\
LAM~\cite{he2025lam} & 16.354 & 0.759 & 0.251 & 0.608 & 5.772 & 0.516 & 5.079 \\
CAP4D~\cite{taubner2025cap4d} & 19.295 & 0.811 & \underline{0.195} & 0.719 & 6.561 & 0.655 & 5.064 \\
\midrule
Our VGM & \textbf{21.586} & \textbf{0.831} & \textbf{0.174} & \textbf{0.754} & \textbf{7.127} & \textbf{0.671} & 5.066 \\
GeoDiff4D & \underline{19.951} & \underline{0.822} & 0.195 & \underline{0.721} & \underline{6.720} & \underline{0.656} & \textbf{5.178} \\
\bottomrule
\end{tabular}%
}
\caption{Quantitative results of self-reenactment on the NeRSemblev2 dataset and cross-reenactment on a mixture of NeRSemblev2 subset and in-the-wild motion data. \textbf{Bold} and \underline{underlined} indicate the best and second-best results, respectively.}
\label{table:metrics}
\end{table}

\subsection{Cross-Reenactment}
We conduct cross-identity animation experiments to evaluate performance across diverse subjects. Following Sec.~\ref{sec:seflreenact}, reference images are sampled from 10 unseen NeRSemblev2 identities and paired with driving sequences from different subjects. To test generalization to extreme expressions and poses, we additionally include an in-the-wild collection comprising 10 reference identities (real humans and cartoon characters) and 6 driving sequences. Driving sequences are randomly sampled from a mixture of NeRSemblev2 and in-the-wild motion data, covering a wide range of expressions and motions.

\begin{figure*}[ht]
  \centering
  \includegraphics[width=0.9\linewidth]{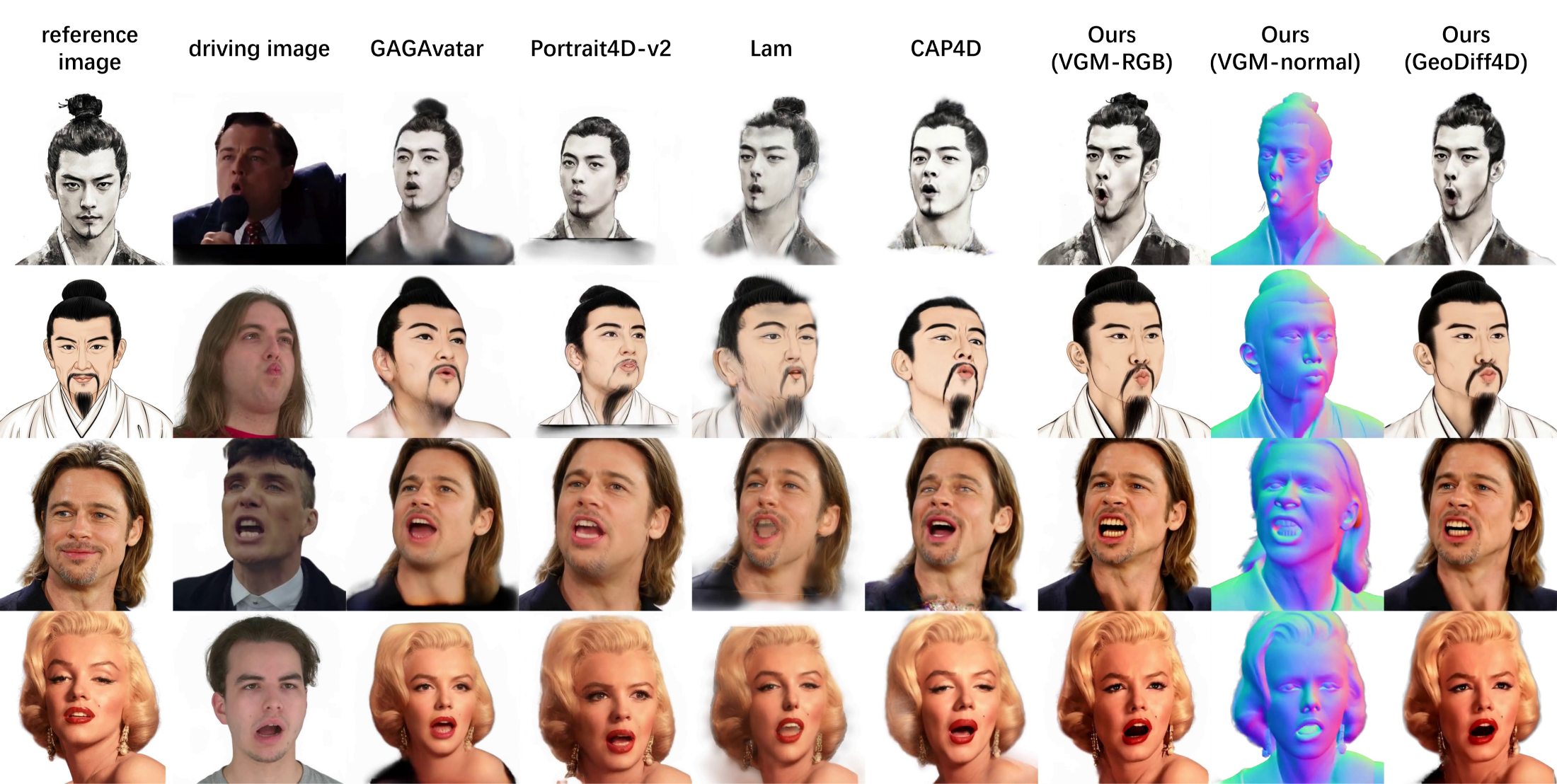}
  \caption{Cross-Reenactment results. Evaluated on a mixture of NeRSemblev2 and in-the-wild collections spanning diverse real and cartoon identities to assess model generalizability.}
  \label{fig:crossreenact}
\end{figure*}

As reported in Table~\ref{table:metrics}, while GeoDiff4D and VGM may not top every quantitative metric, further experiments in Figure~\ref{fig:crossreenact} show they deliver superior visual quality under extreme head poses and exaggerated expressions. Additional results are provided in supplementary material.

\subsection{Ablation Study and Extensions}
% vgm: joint representation
% vgm: cross view pairing
% gs: normal reg
% gs: refinements
\paragraph{Ablation Study.}
We conduct ablation studies on the NeRSemblev2 self-reenactment set to evaluate the contribution of key components. For the video generation model, we ablate joint representation learning, the Domain-Spatial attention module, cross-view pairing, and synthetic data. For avatar reconstruction, we ablate hierarchical refinement, surface normal regularization, and the source of normal supervision (generated vs. monocular estimated). Qualitative results are provided in the supplement.

Quantitative results are shown in Tab.~\ref{table:ablation}. Our full VGM achieves the best overall performance across nearly all metrics. Removing joint representation learning degrades geometry awareness, causing notable drops in reconstruction quality. Without domain attention, information exchange between portrait and geometric representations is weakened, leading to consistent degradation. Cross-view pairing proves most critical, as its removal causes the largest overall performance drop, highlighting the importance of multi-view geometric constraints. Training without synthetic data limits identity diversity and reduces generalization.

For head avatar reconstruction, removing either the hierarchical refinement or normal regularization degrades performance across various metrics. Replacing video-generated normals with monocular estimated normals similarly shows marginal quantitative differences. However, qualitative results in the supplementary material reveal more pronounced distinctions across all ablated variants: the complete model produces fewer artifacts and superior temporal stability, while video-generated normals further contribute to finer facial detail, benefiting from their fine-grained, temporally coherent nature and better alignment with RGB inputs.

\begin{table}[ht]
\centering
\resizebox{\columnwidth}{!}{%
\begin{tabular}{@{}llccccccc@{}}
\toprule
\textbf{Method} & \textbf{Ablation} & PSNR $\uparrow$ & SSIM $\uparrow$ & LPIPS $\downarrow$ & CSIM $\uparrow$ & JOD $\uparrow$ & AKD $\downarrow$ & AED $\downarrow$ \\
\midrule
\multirow{5}{*}{\textbf{VGM only}} &
w/o joint rep. & 20.809 & 0.825 & 0.184 & \textbf{0.757} & 6.960 & 4.216 & 2.489 \\
& w/o domain attn. & 20.984 & 0.820 & 0.179 & 0.743 & 7.029  & 4.195 & 2.556 \\
& w/o cv pairing & 19.895 & 0.804 & 0.191 & 0.734 & 6.859 & 5.367 & 3.113 \\
& w/o synt. data & 20.892 & 0.820 & 0.181 & 0.743 & 6.978 & 4.339 & 2.527 \\
& Ours & \textbf{21.586} & \textbf{0.831} & \textbf{0.174} & 0.754 & \textbf{7.127} & \textbf{4.016} & \textbf{2.340} \\
\midrule
\multirow{3}{*}{\textbf{Recon}} &
w/o hier. refine & 19.816 & 0.821 & 0.198 & 0.736 & 6.758 & \textbf{4.227} & 2.603 \\
& w/o norm. reg & 19.950 & 0.821 & 0.196 & 0.734 & 6.774 & 4.291 & 2.713 \\
& w DAViD norm. & 19.947 & 0.822 & 0.196 & 0.736 & \textbf{6.782} & 4.247 & \textbf{2.553} \\
& Ours & \textbf{19.953} & \textbf{0.822} & \textbf{0.195} & \textbf{0.737} & 6.780 & 4.248 & 2.563 \\
\bottomrule
\end{tabular}%
}
\caption{Ablation study. For video generation model, we ablate joint representation, Domain-Spatial attention module, cross-view pairing strategy and synthetic data. For avatar reconstruction, we ablate hierarchical refinement, normal regularization and generated normal(vs. normal from DAViD).}
\label{table:ablation}
\end{table}

\paragraph{Extensions.}
\label{experiment:expencoder}
% We evaluate the view consistency of our pose-free expression encoder by generating images with expression sequences from 12 different camera viewpoints while maintaining a fixed frontal head pose. This isolates the encoder's ability to capture facial expressions independently of head orientation. Results demonstrate strong pose disentanglement with consistent and accurate expressions across all viewpoints. Qualitative results are provided in the supplementary material.
We validate several additional aspects of our framework: (1) view-consistency of the expression encoder, (2) the quality gap between synthetic and monocular-estimated normals and its benefit to normal prediction, (3) the contributions of hierarchical refinement and normal regularization to free-view synthesis, and (4) the quality-efficiency trade-off across different diffusion sampling steps. Please refer to the supplementary material for full results and comparisons.
\section{Disscusion and Conclusion}
\label{sec:conclusion}
\subsection{Limitations}
% While our method produces compelling 3D head avatars from a single image, several limitations remain. First, we rely heavily on monocular 3DMM tracking for accurate head pose estimation, which is inherently challenging due to its ill-posed nature. Second, although our video generation model supports tongue motion, the final avatar cannot accurately reconstruct tongue movements due to FLAME's structural limitations. Finally, like other diffusion-based methods, our approach suffers from relatively slow sampling times compared to feedforward alternatives. Addressing these limitations and balancing reconstruction quality with inference efficiency represent important directions for future work.
While our method produces compelling 3D head avatars from a single image, several limitations remain. First, we rely heavily on monocular 3DMM tracking for head pose estimation, which is inherently challenging due to its ill-posed nature. Second, although our video generation model supports tongue motion, the final avatar cannot accurately reconstruct tongue movements due to FLAME's structural limitations and lack of fine-grained tongue parameterization. Finally, like other diffusion-based methods, our approach suffers from relatively slow sampling times compared to feedforward alternatives, hindering real-time deployment. Addressing these limitations and improving the balance between reconstruction quality and inference efficiency are important directions for future work.

\subsection{Conclusion}
We present GeoDiff4D, a novel framework for high-fidelity 4D head avatar reconstruction from a single portrait image. By integrating a pose-free expression encoder, a joint appearance-geometry diffusion model, and 3D Gaussian Splatting-based reconstruction, our method generates photorealistic and animatable avatars with strong geometry priors and consistent identity-expression details. Extensive experiments demonstrate superior performance in identity preservation, expression fidelity, and cross-view consistency across diverse subjects and challenging poses. Our approach effectively bridges diffusion-based generation and 3D avatar reconstruction, advancing high-quality digital human creation.

\noindent\textbf{Social Impact.} 
As with other generative avatar technologies, our approach may introduce risks related to identity misuse. We advocate for responsible deployment, including consent-aware data use and provenance tracking.

\section*{Acknowledgments}
This work was supported by the National Natural Science Foundation of China (NSFC) under Grant 62125107.
{
    \small
    \bibliographystyle{ieeenat_fullname}
    \bibliography{main}
}

% % WARNING: do not forget to delete the supplementary pages from your submission 
% \clearpage
% \setcounter{page}{1}
% \maketitlesupplementary

% \section{Rationale}
% \label{sec:supp}
% % 
% Having the supplementary compiled together with the main paper means that:
% % 
% \begin{itemize}
% \item The supplementary can back-reference sections of the main paper, for example, we can refer to \cref{sec:intro};
% \item The main paper can forward reference sub-sections within the supplementary explicitly (e.g. referring to a particular experiment); 
% \item When submitted to arXiv, the supplementary will already included at the end of the paper.
% \end{itemize}
% % 
% To split the supplementary pages from the main paper, you can use \href{https://support.apple.com/en-ca/guide/preview/prvw11793/mac#:~:text=Delete%20a%20page%20from%20a,or%20choose%20Edit%20%3E%20Delete).}{Preview (on macOS)}, \href{https://www.adobe.com/acrobat/how-to/delete-pages-from-pdf.html#:~:text=Choose%20%E2%80%9CTools%E2%80%9D%20%3E%20%E2%80%9COrganize,or%20pages%20from%20the%20file.}{Adobe Acrobat} (on all OSs), as well as \href{https://superuser.com/questions/517986/is-it-possible-to-delete-some-pages-of-a-pdf-document}{command line tools}.

\clearpage
\setcounter{page}{1}
\maketitlesupplementary

\setcounter{section}{0}
\renewcommand{\thesection}{\Alph{section}}

\definecolor{best}{RGB}{250,227,127}
\definecolor{second}{RGB}{237,186,145}

\newcommand{\hlbest}[1]{\setlength{\fboxsep}{1pt}\colorbox{best}{#1}}
\newcommand{\hlsecond}[1]{\setlength{\fboxsep}{0.5pt}\colorbox{second}{#1}}

\section{Implementation Details}

\subsection{Video Generation Model}
\paragraph{Noise Schedule}
During inference, we do not initialize the diffusion process from pure random noise. Instead, we inject reference information directly into the initial latent by adding Gaussian noise to both the reference image and its corresponding normal map, and then use these noisy signals as the starting point for DDIM denoising. This reference-conditioned initialization encourages the model to preserve identity features, fine-grained facial details, and geometric structure more faithfully throughout the iterative sampling process, reducing ambiguity compared to fully noise-based initialization. Concretely, we first construct the two domain-specific latents (image and normal) independently and then concatenate them along the domain dimension to form an initial latent of shape $[B \times D \times C \times T \times H \times W]$, following the formulation described in Section~3. By starting from this structurally coherent latent, the model benefits from stronger reference guidance, resulting in more consistent and identity-preserving outputs across both spatial and temporal dimensions.

\paragraph{Classifier-Free Guidance}
During training, we randomly drop the expression latent with a probability of 0.1 to facilitate classifier-free guidance, enabling the model to learn to generate outputs with and without explicit expression conditioning. During inference, classifier-free guidance is applied simultaneously to both domains to maintain consistent control over the generated image and normal signals. Specifically, after constructing the initial noisy latents as described previously, we first reshape them into an image–normal latent of shape $[(B \times D) \times C \times T \times H \times W]$. This latent is then duplicated along the batch–domain dimension to form an unconditional–conditional latent pair, which allows the guidance mechanism to differentiate between conditioned and unconditioned signals. All other conditioning inputs, including expression latents, head pose maps, and class labels, are processed in the same manner to maintain alignment across all control signals. Finally, we apply the standard classifier-free guidance procedure with a guidance scale of 2.5 for all experiments, ensuring strong adherence to the desired expressions and head poses while enhancing overall fidelity and consistency across both spatial and temporal dimensions.

\paragraph{Video Synthesis for Reconstruction}
We use the video generation model to synthesize videos of approximately 200 frames across 12 viewpoints. Specifically, we select 12 out of the 16 available viewpoints from NeRSemblev2 and adopt their camera configuration, keeping the head pose fixed while animating only the facial expressions of the reference image. We then concatenate all video clips into a single sequence, treating it as a monocular video, which is subsequently used for avatar reconstruction.
% We pre-construct an expression sequence and a head-pose sequence to serve as control signals for the diffusion model, enabling the generation of videos from any given reference image. These videos are subsequently used for head avatar reconstruction. The expression sequence is drawn from the NeRSemblev2 dataset and an in-the-wild collection described in Section 5, covering a wide range of rich and exaggerated facial expressions. All clips are concatenated into a single continuous sequence of approximately 2,400 frames. For head pose, the FLAME mesh is kept fixed while camera extrinsics are varied to generate smooth interpolations, spanning a yaw range of $\pm 35^\circ$ and a pitch range of $\pm 20^\circ$, ensuring diverse head orientations. By combining these expression and pose sequences, we produce long videos that capture extensive facial dynamics and head rotations, which substantially enhance the robustness and quality of reconstructed head avatars under novel expressions and viewpoints.

\subsection{4D Reconstruction}
\paragraph{FLAME Refinement}
To refine the initial FLAME tracking and compensate for errors from monocular fitting, we introduce learnable residuals for all FLAME parameters, including shape, expression, jaw, neck, eye, and eyelid coefficients, as well as global pose $(R, t)$. These parameters are jointly optimized with the Gaussian attributes using a three-phase learning-rate schedule (warmup, stable, decay). We divide the FLAME parameters into two groups: pose- and shape-related parameters are trained with a conservative schedule and a peak learning rate of $1\mathrm{e}{-5}$, while expression-related parameters adopt a higher peak learning rate of $1\mathrm{e}{-4}$ to better capture fine-grained facial dynamics.

All parameters start from an extremely low learning rate of $1\mathrm{e}{-10}$ and linearly ramp up during the first 40K iterations to prevent instability in early training. The learning rate is maintained at its peak from 20K to 80K iterations to allow sufficient exploration and then exponentially decayed from 80K to 100K iterations to ensure convergence. This staged schedule, combined with group-specific learning rates, enables stable joint optimization of tracking parameters and Gaussian attributes while preserving photometric fidelity and temporal coherence.

\paragraph{Topology-Preserving Remeshing}
To ensure geometric consistency when remeshing the FLAME template into UV space, we adopt a topology-preserving strategy that validates the connectivity of newly generated faces. Standard UV remeshing subdivides each UV grid cell into two triangles to create a dense tessellation, but this approach can inadvertently produce invalid faces that connect distant or topologically unrelated regions of the original mesh. Such invalid connections can lead to visual artifacts and geometric distortions during rendering and deformation.

To prevent such artifacts, we introduce an adjacency-based validation mechanism grounded in the original FLAME topology. We first construct a face-adjacency graph encoding connectivity between all FLAME faces, which is precomputed and cached for efficiency. For each candidate UV face, we retrieve the FLAME face indices of its three vertices through the UV rasterization mapping. A UV face is retained only if (1) all vertices belong to the same FLAME face, or (2) the vertices span multiple FLAME faces that are mutually connected within a bounded hop distance in the adjacency graph. This multi-hop connectivity check is performed via breadth-first search (BFS) with a maximum hop threshold of 5.
This mechanism effectively filters out topologically invalid or distorted triangles while preserving sufficient mesh density for high-quality Gaussian splatting. The resulting UV mesh maintains the original FLAME topology and provides a reliable, deformation-aware surface for attaching Gaussian attributes.

\begin{figure*}[ht]
  \centering
  \includegraphics[width=\linewidth]{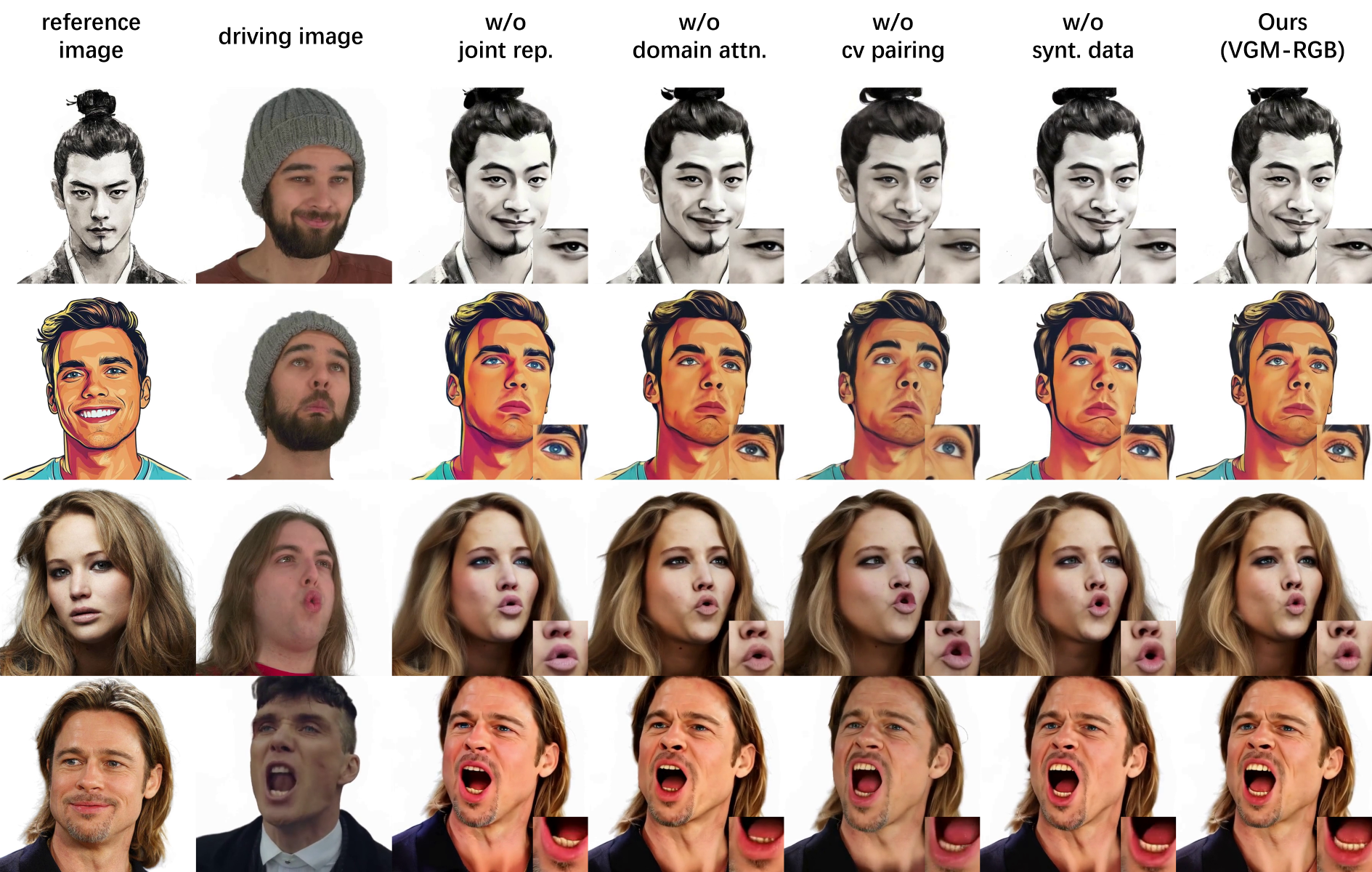}
  \caption{Ablation of portrait generation. We ablate joint representation learning, the Domain-Spatial attention module, cross-view pairing, and synthetic data.}
  \label{fig:ablation-vgm}
\end{figure*}

\section{More Ablation Results}

\subsection{VGM Ablation}
\paragraph{Portrait Generation}
Qualitative results are presented in Fig.~\ref{fig:ablation-vgm}. Our full model shows clear improvements in fine facial details, such as wrinkles, eyelashes, and teeth, compared to the ablated versions. When joint representation learning is removed, the model's ability to transfer expressions is noticeably weakened. We believe this is due to the absence of 3D consistency provided by the surface normal domain, which increases the entanglement between identity, head pose, and expression. Similarly, removing the domain attention module degrades both expression transfer and the fidelity of facial details, emphasizing the importance of cross-domain information exchange between image and normal features. Ablating the cross-view pairing strategy leads to obvious identity leakage and degraded driving quality, where the generated results inappropriately retain characteristics from the driving sequence rather than faithfully following the reference identity and driving expressions. The results obtained without synthetic data are closer to the complete model, indicating that while the inclusion of synthetic data provides additional gains in generation quality, the contributions of the core modules—joint representation, domain attention, and cross-view pairing—are essential for robust and high-fidelity video generation.

\begin{figure*}[ht]
  \centering
  \includegraphics[width=0.8\linewidth]{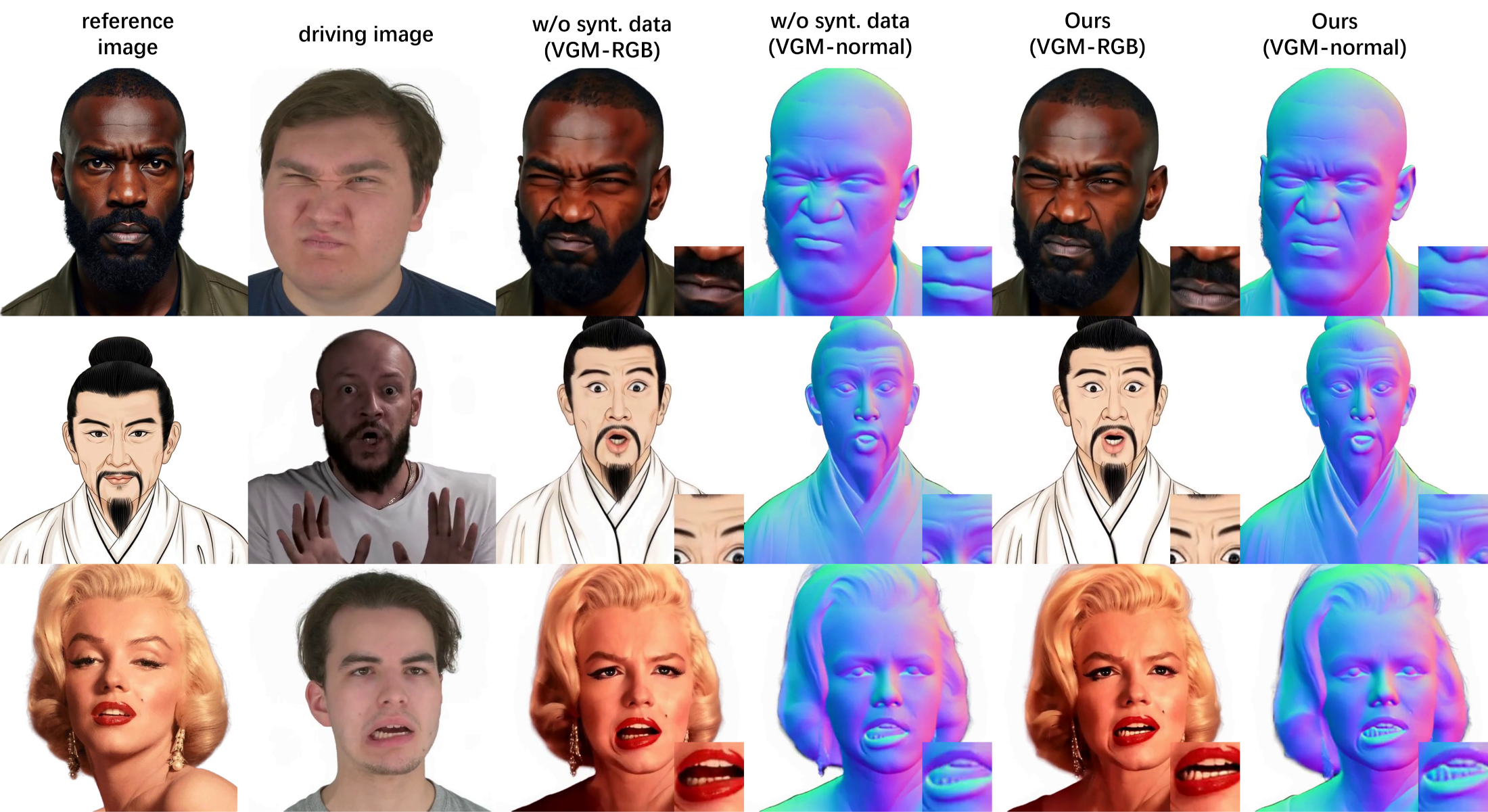}
  \caption{Ablation of normal generation. We ablate synthetic data to evaluate its contribution to normal generation quality.}
  \label{fig:ablation-normal}
\end{figure*}

\begin{figure*}[ht]
  \centering
  \includegraphics[width=\linewidth]{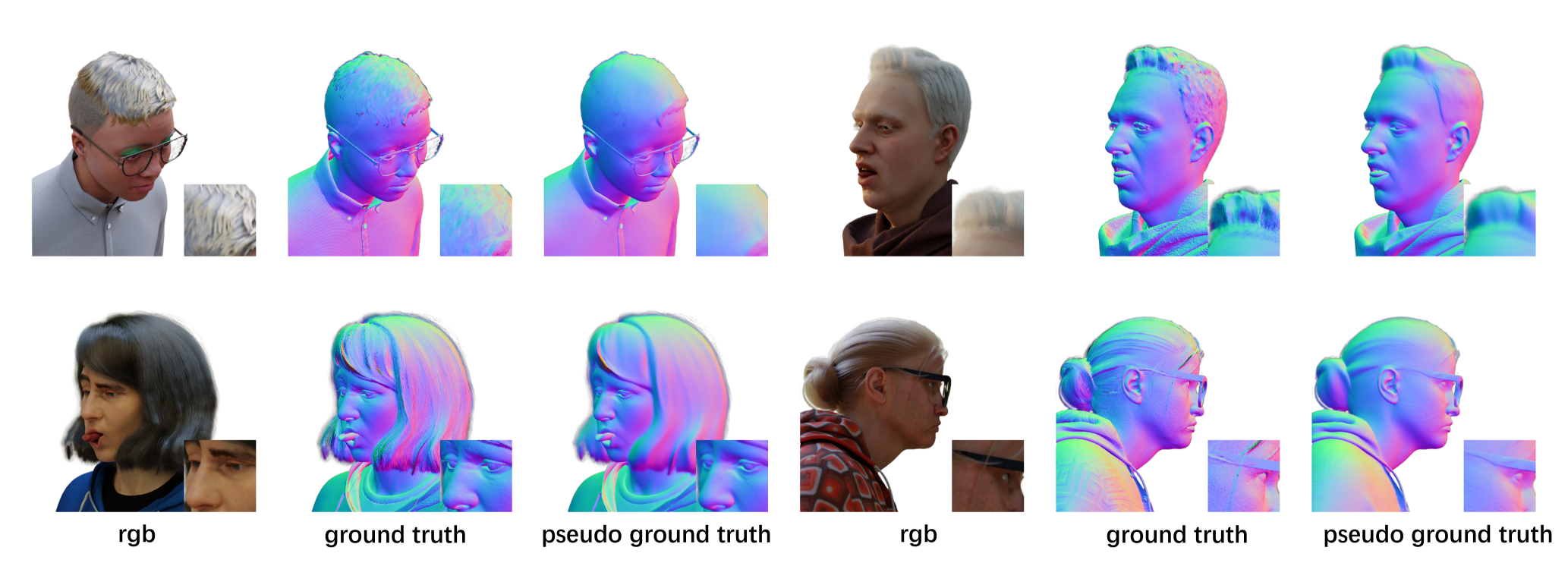}
  \caption{Comparison of ground-truth and pseudo-ground-truth normals. Ground-truth normals from synthetic data are more accurate and capture finer geometric details, thereby improving the quality of generated surface normals.}
  \label{fig:david}
\end{figure*}

\paragraph{Normal Generation}
We further evaluate the impact of synthetic data on the quality of generated surface normals. As shown in Fig.~\ref{fig:ablation-normal}, removing synthetic data results in a modest but noticeable reduction in high-frequency facial details, such as wrinkles and fine contours. Incorporating synthetic data provides an overall improvement in visual fidelity, particularly in regions that are challenging to reconstruct from pseudo-ground-truth normals alone. We believe this improvement is primarily due to the high accuracy and fine-grained detail of the synthetic normals, which offer stronger geometric supervision compared to the limited-fidelity pseudo-ground-truth normals present in other datasets. This effect is further illustrated in Fig.~\ref{fig:david}, where synthetic normals help preserve subtle facial geometry that is otherwise lost.

\subsection{Head Avatar Ablation}
\begin{figure*}[ht]
  \centering
  \includegraphics[width=0.8\linewidth]{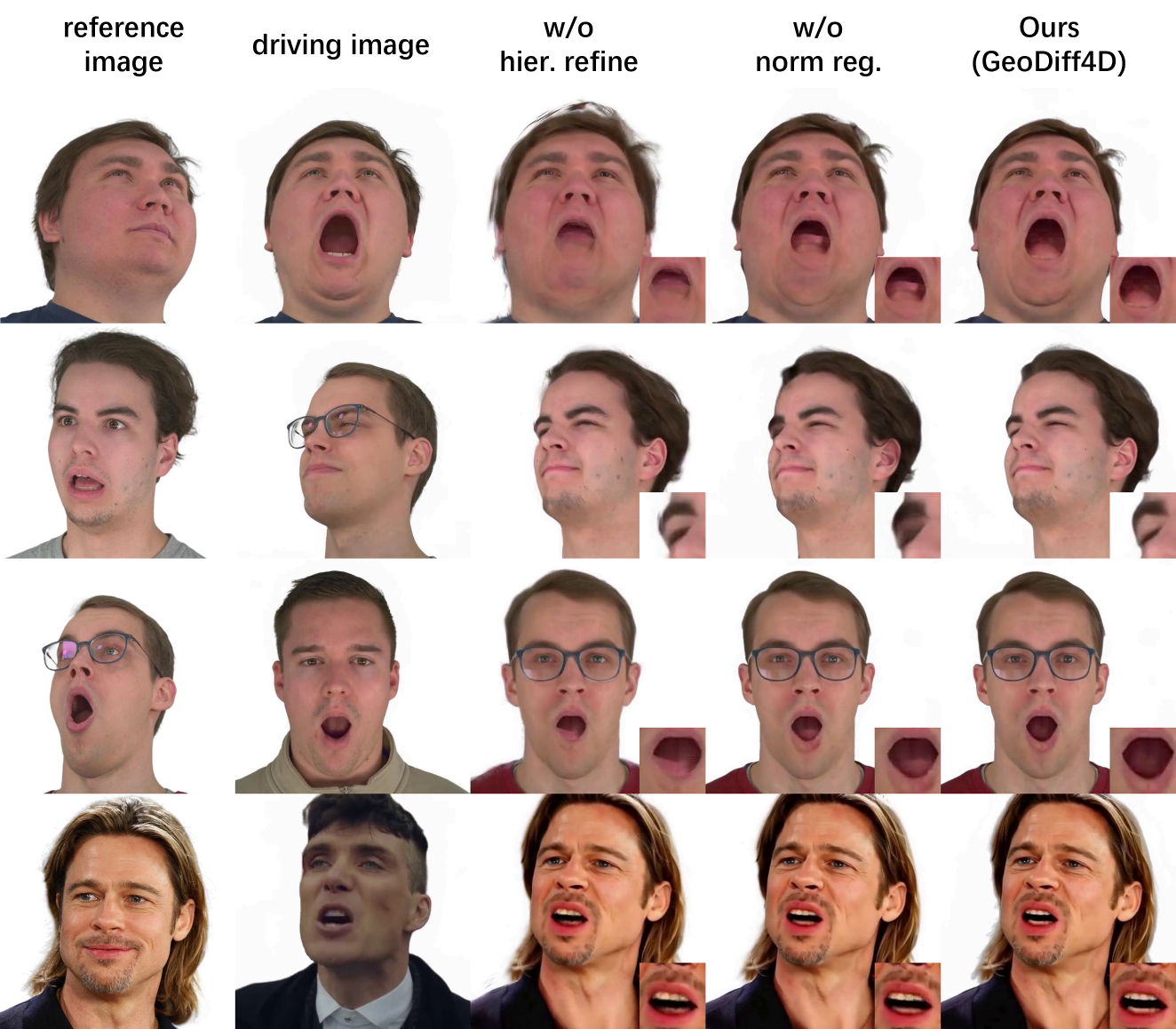}
  \caption{Ablation on 4D reconstruction. We ablate hierarchical refinement and normal regularization to evaluate their contributions to reconstruction quality.}
  \label{fig:ablation-gs}
\end{figure*}

\begin{figure*}[ht]
  \centering
  \includegraphics[width=\linewidth]{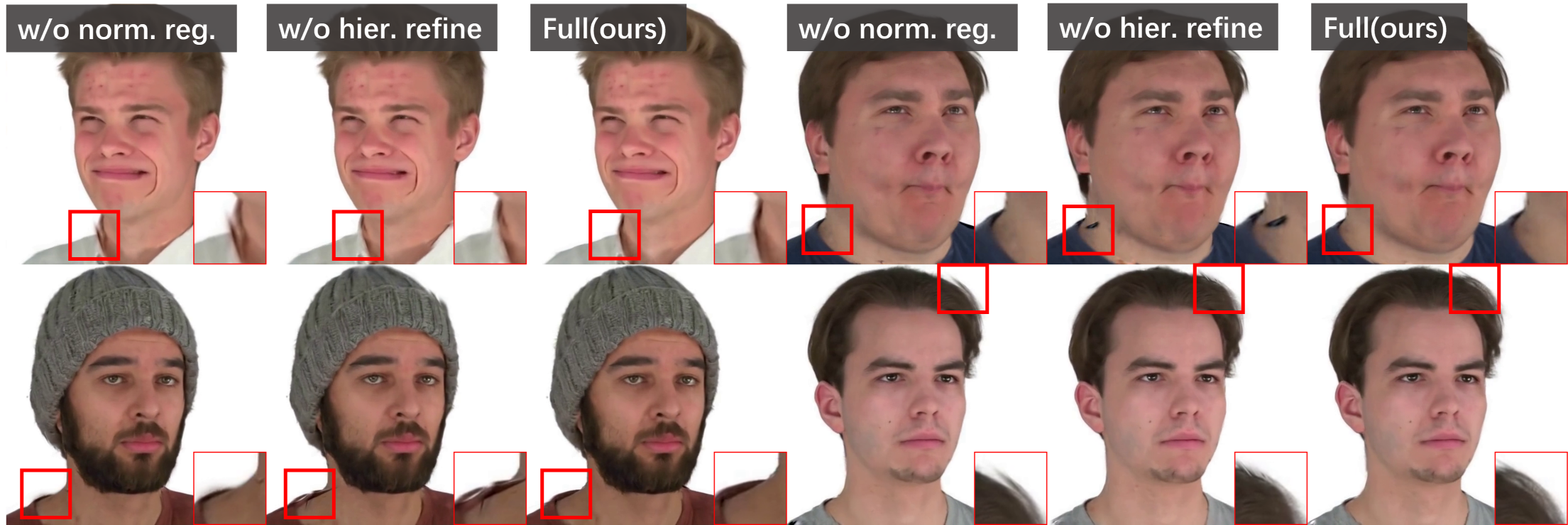}
  \caption{Ablations on hierarchical refinement and normal regularization for free-view rendering.}
  \label{fig:ablation-fv}
\end{figure*}

\begin{figure*}[ht]
  \centering
  \includegraphics[width=\linewidth]{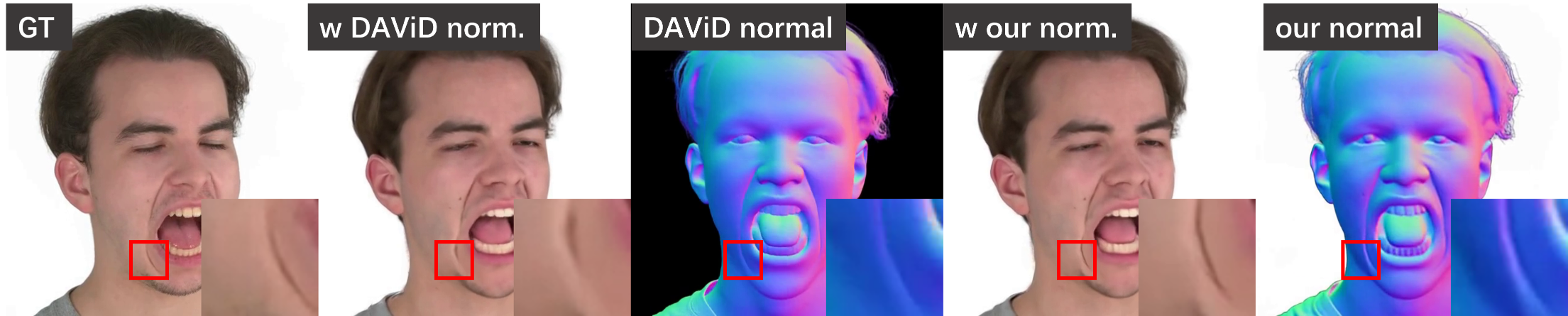}
  \caption{Ablations on our normals versus DAViD normals.}
  \label{fig:ablation-norm}
\end{figure*}

Qualitative results in Fig.~\ref{fig:ablation-gs} show that removing hierarchical refinement leads to noticeably degraded reconstruction quality, particularly in sequences with large head rotations and exaggerated expressions, where errors from monocular FLAME tracking propagate more severely without hierarchical correction. Ablating normal regularization further highlights its critical role in providing geometric guidance for challenging regions such as the mouth and teeth, and in mitigating artifacts under extreme head poses. Fig.~\ref{fig:ablation-fv} further demonstrates the contribution of both modules to free-view rendering quality, where the full model significantly reduces artifacts and produces more faithful novel-view synthesis. As shown in Fig.~\ref{fig:ablation-norm}, normals generated by our model capture finer facial details compared to monocular estimated normals, enabling more accurate geometric guidance for Gaussian splatting optimization. Overall, these results confirm that hierarchical refinement, normal regularization, and high-quality generated normals are all essential for producing high-fidelity 3D head avatars.

\section{More Results}
\subsection{Comparisons with More Baselines}

\begin{table*}[ht]
\centering
\captionsetup{skip=1pt}
\tiny
\setlength{\aboverulesep}{1pt}
\setlength{\belowrulesep}{1pt}
\resizebox{\columnwidth}{!}{%
\begin{tabular}{@{}l*{5}{c}@{}}
\toprule
\textbf{Method} & 
PSNR $\uparrow$ & SSIM $\uparrow$ & LPIPS $\downarrow$ & CSIM $\uparrow$ & JOD $\uparrow$ \\
\midrule
VOODOOXP & 13.111 & 0.714 & 0.276 & 0.598 & 5.116 \\
LivePortrait & 18.047 & 0.787 & 0.214 & 0.736 & 6.230 \\
Wan-Animate & 12.805 & 0.655 & 0.354 & 0.712 & 4.597 \\
X-NeMo & 14.202 & 0.683 & 0.311 & 0.732 & 4.987 \\
\midrule
Our VGM & \hlbest{21.586} & \hlbest{0.831} & \hlbest{0.174} & \hlbest{0.754} & \hlbest{7.127} \\
GeoDiff4D & \hlsecond{19.953} & \hlsecond{0.822} & \hlsecond{0.195} & \hlsecond{0.737} & \hlsecond{6.780} \\
\bottomrule
\end{tabular}%
}
\caption{Quantitative comparison with more baselines (\hlbest{best}, \hlsecond{second-best}).}
\label{table:moreresults}
\end{table*}

\begin{figure*}[ht]
  \centering
  \includegraphics[width=\linewidth]{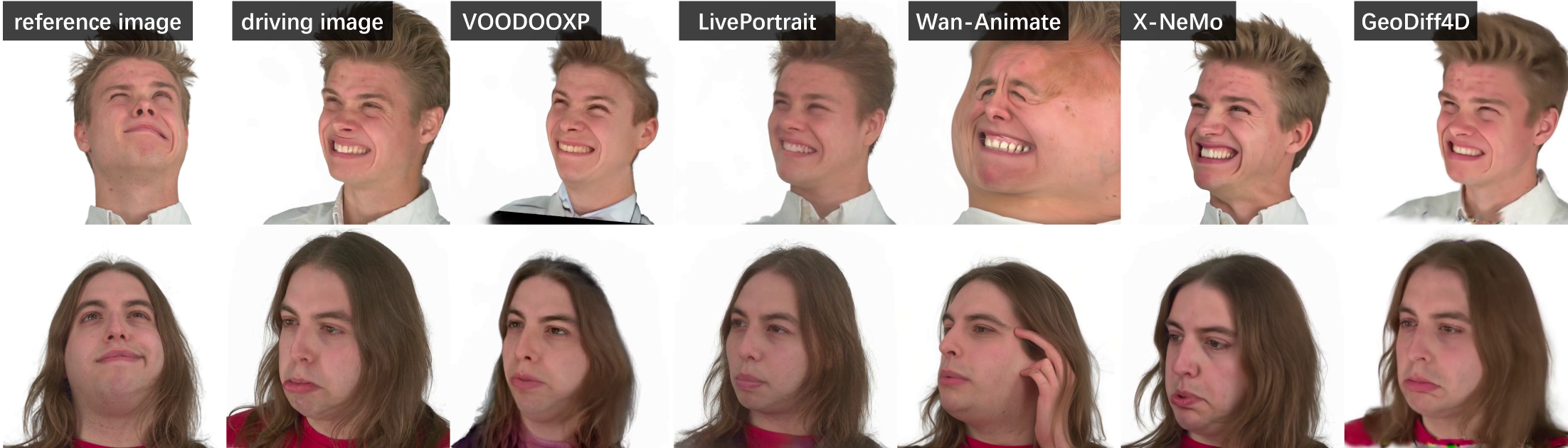}
  \caption{Qualitative comparison with more baselines.}
  \label{fig:moreresults}
\end{figure*}

% We provide additional comparisons with more baselines in Tab.~\ref{table:moreresults} and Fig.~\ref{fig:moreresults}, including diffusion-based generative models X-NeMo~\cite{zhao2025x} and Wan-Animate~\cite{wan2025}, as well as LivePortrait~\cite{guo2024liveportrait} and VoodooXP~\cite{tran2024voodooxpexpressiveoneshot}. Both qualitative and quantitative results show that our method achieves the best performance on most metrics, delivering more accurate expression transfer and remaining robust under large head pose variations.

We provide additional comparisons with more baselines in Tab.~\ref{table:moreresults} and Fig.~\ref{fig:moreresults}, including diffusion-based generative models X-NeMo and Wan-Animate, as well as LivePortrait and VoodooXP. Both qualitative and quantitative results show that our method achieves the best performance on most metrics, delivering more accurate expression transfer and remaining robust under large head pose variations.

\begin{figure*}[ht]
  \centering
  \includegraphics[width=\linewidth]{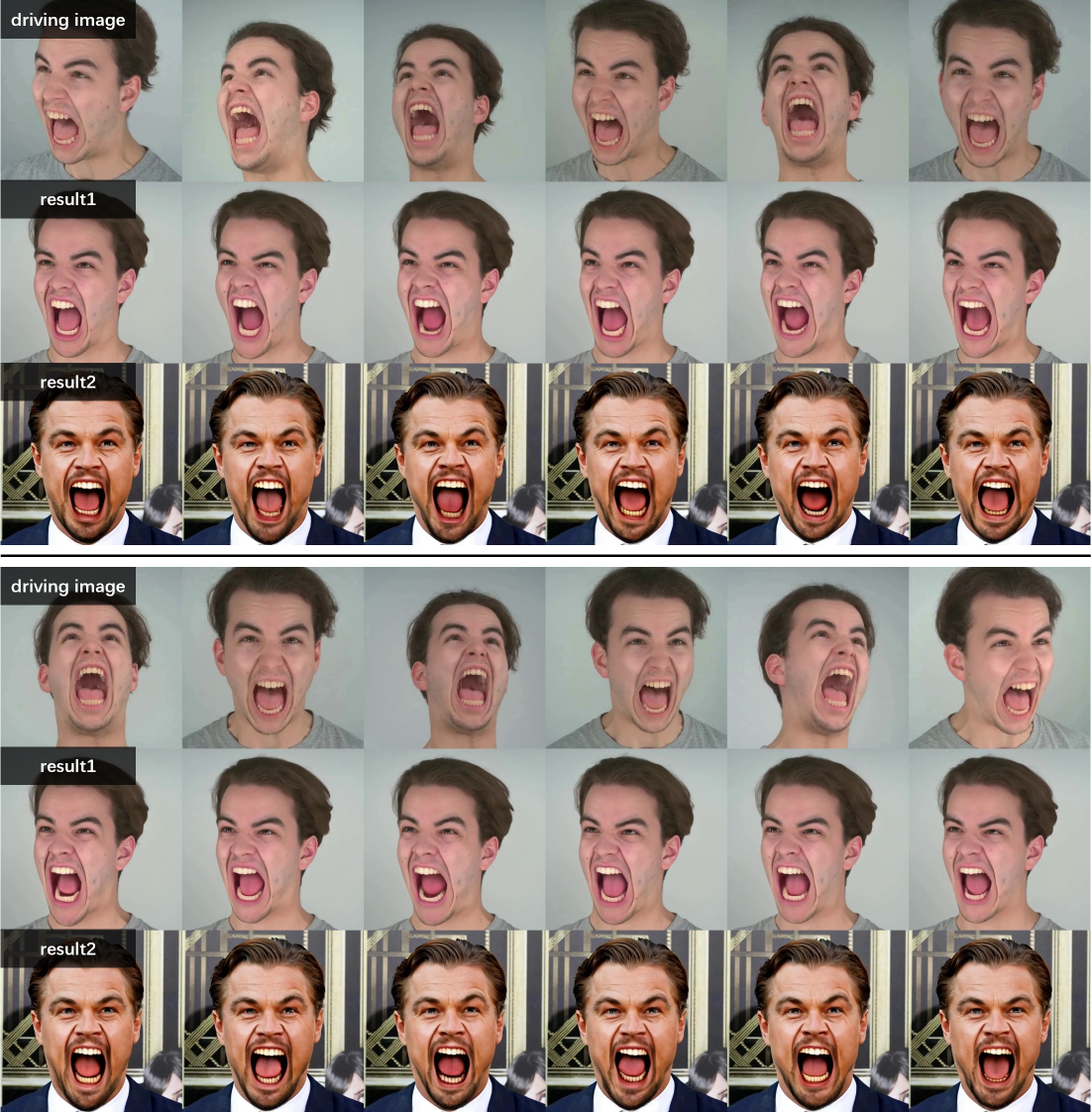}
  \caption{Cross-view animation results. We generate images using expression sequences from 12 different camera viewpoints while fixing the head pose, demonstrating the view consistency of our pose-free expression encoder.}
  \label{fig:view}
\end{figure*}

\subsection{Cross-View Animation}
The results illustrated in Fig.~\ref{fig:view} demonstrate that our pose-free expression encoder exhibits strong cross-view consistency, producing highly consistent facial expressions across a wide range of camera viewpoints. Even when the head pose is held fixed, the encoder effectively disentangles expression from pose, maintaining detailed and coherent facial dynamics regardless of the viewing angle. This highlights the robustness of our encoder in preserving expression fidelity across diverse perspectives.

\subsection{Computational Efficiency Discussion}
As computational efficiency is a practical concern for diffusion-based models, we evaluate generation quality against inference cost across different sampling steps (Tab.~\ref{table:sampling-steps}). We adopt 25 steps as our default, striking a favorable balance between quality and speed.

\begin{table*}[ht]
\centering
\captionsetup{skip=1pt}
\setlength{\aboverulesep}{1pt}
\setlength{\belowrulesep}{1pt}
\resizebox{\columnwidth}{!}{%
\begin{tabular}{@{}c*{8}{c}@{}}
\toprule
\textbf{Steps} & PSNR $\uparrow$ & SSIM $\uparrow$ & LPIPS $\downarrow$ & CSIM $\uparrow$ & JOD $\uparrow$ & AKD $\downarrow$ & AED $\downarrow$ & Speed (s/frame) $\downarrow$ \\
\midrule
10  & 21.177 & \hlsecond{0.8286} & 0.1779 & \hlsecond{0.7419} & 7.033 & \hlbest{4.215} & \hlbest{2.503} & \hlbest{1.16} \\
25  & 21.505 & \hlbest{0.8294} & 0.1746 & \hlsecond{0.7419} & 7.093 & \hlsecond{4.242} & \hlsecond{2.541} & \hlsecond{2.74} \\
50  & \hlsecond{21.529} & 0.8285 & \hlsecond{0.1740} & \hlbest{0.7421} & \hlbest{7.097} & 4.283 & 2.573 & 5.39 \\
100 & \hlbest{21.545} & 0.8280 & \hlbest{0.1736} & 0.7412 & \hlbest{7.100} & 4.261 & 2.550 & 10.66 \\
\bottomrule
\end{tabular}%
}
\caption{Ablation on sampling steps (\hlbest{best}, \hlsecond{second-best}).}
\label{table:sampling-steps}
\end{table*}

\end{document}